\documentclass[twocolumn]{article}
\usepackage{geometry}
\usepackage[utf8]{inputenc}
\usepackage{graphicx}
\usepackage{amsmath}
\usepackage{mathtools}
\usepackage{centernot}
\usepackage{amssymb}
\usepackage{hyperref}
\usepackage{microtype}
\usepackage{cleveref}
\hypersetup{colorlinks=true, allcolors=[rgb]{0.3, 0.3, 0.3}}

\usepackage{natbib}
\usepackage{bm}
\usepackage[retainorgcmds]{IEEEtrantools}
\usepackage[textsize=tiny, textwidth=60mm, disable]{todonotes}
\usepackage[aboveskip=1pt,labelfont=bf,labelsep=period,singlelinecheck=off,font=small]{caption}

\newcommand{\E}{ \mathbb{E} }
\newcommand{\Prob}{ P }
\newcommand{\giv}{ \,|\, }

\newcommand\indep{\protect\mathpalette{\protect\independenT}{\perp}}
\def\independenT#1#2{\mathrel{\rlap{$#1#2$}\mkern3.5mu{#1#2}}}
\newcommand\nindep{\protect\centernot\indep}

\newcommand{\eg}{e.g.\,}
\newcommand{\ie}{i.e.\,}
\newcommand{\aka}{a.k.a.\,}
\newcommand{\etc}{\emph{etc.}\,}

\title{Preventing dataset shift from breaking machine-learning
biomarkers}
\hypersetup{
 pdfauthor={jerome dockes},
 pdftitle={Preventing dataset shift from breaking machine-learning
biomarkers},
 pdfkeywords={},
 pdfsubject={},
 pdflang={English}}

\author{Jérôme Dockès$^{1*}$, Gaël Varoquaux$^{12+}$, Jean-Baptiste Poline$^{1+}$}
\date{$^1$McGill University $^2$INRIA \\ $^*$Corresponding author $^+$JB Poline and Gaël Varoquaux contributed equally to this work.  \vspace{20pt} \\ July 2021}

\begin{document}

\maketitle

\begin{abstract}
  Machine learning brings the hope of finding new biomarkers extracted
  from cohorts with rich biomedical measurements. A good biomarker is one
  that gives reliable detection of the corresponding condition.
  However, biomarkers are often extracted from a cohort that differs
  from the target population. Such a
  mismatch, known as a dataset shift, can undermine the application of the
  biomarker to new individuals.
  Dataset shifts are frequent in biomedical research, \eg because of recruitment biases.
  When a dataset shift occurs, standard machine-learning techniques do not
  suffice to extract and validate biomarkers. This article provides an overview
  of when and how dataset shifts breaks machine-learning extracted
  biomarkers, as well as detection and correction strategies.
\end{abstract}

\section{Introduction: dataset shift breaks learned biomarkers}
Biomarkers are measurements that provide information about a medical condition
or physiological state \citep{strimbu2010biomarkers}. For example, the presence
of an antibody may indicate an infection; a complex combination of features
extracted from a medical image can help assess the evolution of a tumor.
Biomarkers are important for diagnosis, prognosis, and treatment
or risk assessments.

Complex biomedical measures may carry precious medical information,
as with histopathological images or genome sequencing of biopsy samples in
oncology. Identifying quantitative biomarkers from these requires
sophisticated statistical analysis. With large datasets becoming
accessible, supervised machine learning provides new promises by
optimizing the information extracted to relate to a specific output variable of
interest, such as a cancer diagnosis
\citep{andreu2015big,faust2018deep,deo2015machine}. These methods,
cornerstones of artificial intelligence, are starting to
appear in clinical practice: a machine-learning based radiological tool
for breast-cancer diagnosis has recently been approved by the
FDA\footnote{\url{https://fda.report/PMN/K192854}}.

Can such predictive biomarkers, extracted through complex data processing, be safely
used in clinical practice, beyond the initial research settings? One risk
is the potential mismatch, or \emph{dataset shift}, between the distribution
of the individuals used to estimate this statistical link and that of the target
population that should benefit from the biomarker. In this case,
the extracted associations may not apply to the target
population \citep{kakarmath2020best}.
Computer aided diagnostic of thoracic diseases
from X-ray images has indeed been shown to be unreliable for individuals of a 
given sex if built from a cohort over-representing the other sex
\citep{larrazabal2020gender}.
More generally, machine-learning systems may fail on data from different
imaging devices, hospitals, populations with a different age distribution, \etc.
Dataset biases are in fact frequent in medicine. For instance selection
biases --\emph{eg} due to volunteering self-selection, non-response,
dropout...-- \citep{rothman2012epidemiology,tripepi2010selection}
may cause cohorts to capture only a small range of possible patients and
disease manifestations in the presence of 
spectrum effects \citep{ransohoff1978problems,mulherin2002spectrum}.
Dataset shift or dataset bias can
cause systematic errors that cannot be fixed by
acquiring larger datasets and require specific methodological care.

In this article, we consider predictive biomarkers identified with supervised machine learning.
We characterize the problem of dataset shift, show how it can hinder the use
of machine learning for health applications
\citep{woo2017building,wynants2020prediction}, and provide mitigation
strategies.

\section{A primer on machine learning for biomarkers}

\subsection{Empirical Risk Minimization}

Let us first introduce the principles of machine learning used to identify biomarkers.
Supervised learning captures, from observed data,
the link between a set of input
measures (features) $X$ and an output (\eg a condition) $Y$: for example the relation between the absorption spectrum of
oral mucosa and blood glucose concentration \citep{kasahara2018noninvasive}. A
supervised learning algorithm finds a function $f$ such that $f(X)$ is as close as possible to
the output $Y$.
Following machine-learning terminology, we call the system's best guess $f(X)$
for a value $X$ a \emph{prediction}, even when it does not concern a measurement
in the future.

Empirical Risk Minimization, central to machine learning,
uses a loss function $L$ to measure how far a
prediction $f(X)$ is from the true value $Y$, for example the squared
difference:
\begin{equation}
  L(Y, f(X)) = (Y - f(X))^2 \; .
\end{equation}
The goal is to find a function $f$ that has
 a small \emph{risk}, which is the \emph{expected} loss on
the true distribution of \(X\) and \(Y\), \ie on \emph{unseen individuals}.
The true risk cannot be computed in practice: it would require having seen all
possible patients, the true distribution of patients.
The \emph{empirical} risk is used instead: the average error over available
examples,
\begin{equation}
  \hat{R}(f) = \frac{1}{n}\sum_{i=1}^n L(y_i, f(x_i)) \; ,
\end{equation}
where $\{(x_i, y_i)\,,\,i=1,\dots,n\}$ are available $(X, Y)$ data,
called \emph{training} examples.
The statistical link of interest is then approximated by choosing $f$
 within a family of candidate functions as
the one that minimizes the empirical risk \(\hat{R}(f)\).

The crucial assumption underlying this very popular approach is that the
prediction function $f$ will then be applied to individuals drawn from the same
population as the training examples $\{x_i, y_i\}$. It can be important
to distinguish the \emph{source} data, used to fit and evaluate a machine-learning model (e.g. a dataset
collected for research), from the \emph{target} data, on which
predictions are meant to be used for clinical applications (e.g. new visitors of a hospital).
Indeed, if the training
examples are not representative of the target population -- if there is a
dataset shift -- the empirical risk is a poor estimate of the expected error,
and $f$ will not perform well on individuals from the target population.

\subsection{Evaluation: Independent test set and cross-validation}
\label{sec:org87b9cec}
Once a model has been estimated from training examples, measuring its error
on these same individuals results in a (sometimes wildly) optimistic estimate of
the expected error on \emph{unseen} individuals (\citet[Sec.
7.4]{friedman2001elements}, \citet[Sec. 1, ``Association vs Prediction'']{poldrack2020establishment}).
Indeed, predictors chosen from a rich family of functions are very
flexible and can learn rules that fit tightly the training examples but fail
to generalize to new individuals. This is called \emph{overfitting}.

To obtain valid estimates of the expected performance on new data, the
error is measured on an independent sample held out during training, called the
test set.
The most common approach to obtain such a test set is to randomly split the
available data.
This process is usually repeated with several splits, a procedure called
cross-validation \citep[Sec. 7]{arlot2010survey,friedman2001elements}.

When training and test examples are chosen uniformly from the same sample, they
are drawn from the same distribution (i.e. the same population): there is no
dataset shift.
Some studies also measure the error on an \emph{independent} dataset
\citep[e.g.][]{beck2011systematic,jin2020generalizable}. This helps
establishing external validity, assessing
whether the predictor will perform well outside of the dataset used to
define it \citep{bleeker2003external}.
Unfortunately, the biases in participant recruitment may be similar
in independently collected datasets. For example if patients with severe
symptoms are difficult to recruit, this is likely to distort all datasets
similarly. Testing on a dataset collected independently is therefore a useful
check, but no silver bullet to rule out dataset shift issues.

\section{False solutions to tackling dataset shift}\label{sec:misconceptions}

We now discuss some misconceptions and
confusions with problems not directly related to dataset shift.

\begin{figure*}[t!]
  \centering
  \begin{minipage}{.44\textwidth}
    \centerline{\sffamily\bfseries Data generation}

    \hspace*{.05\linewidth}%
    \includegraphics[width=.6\textwidth]{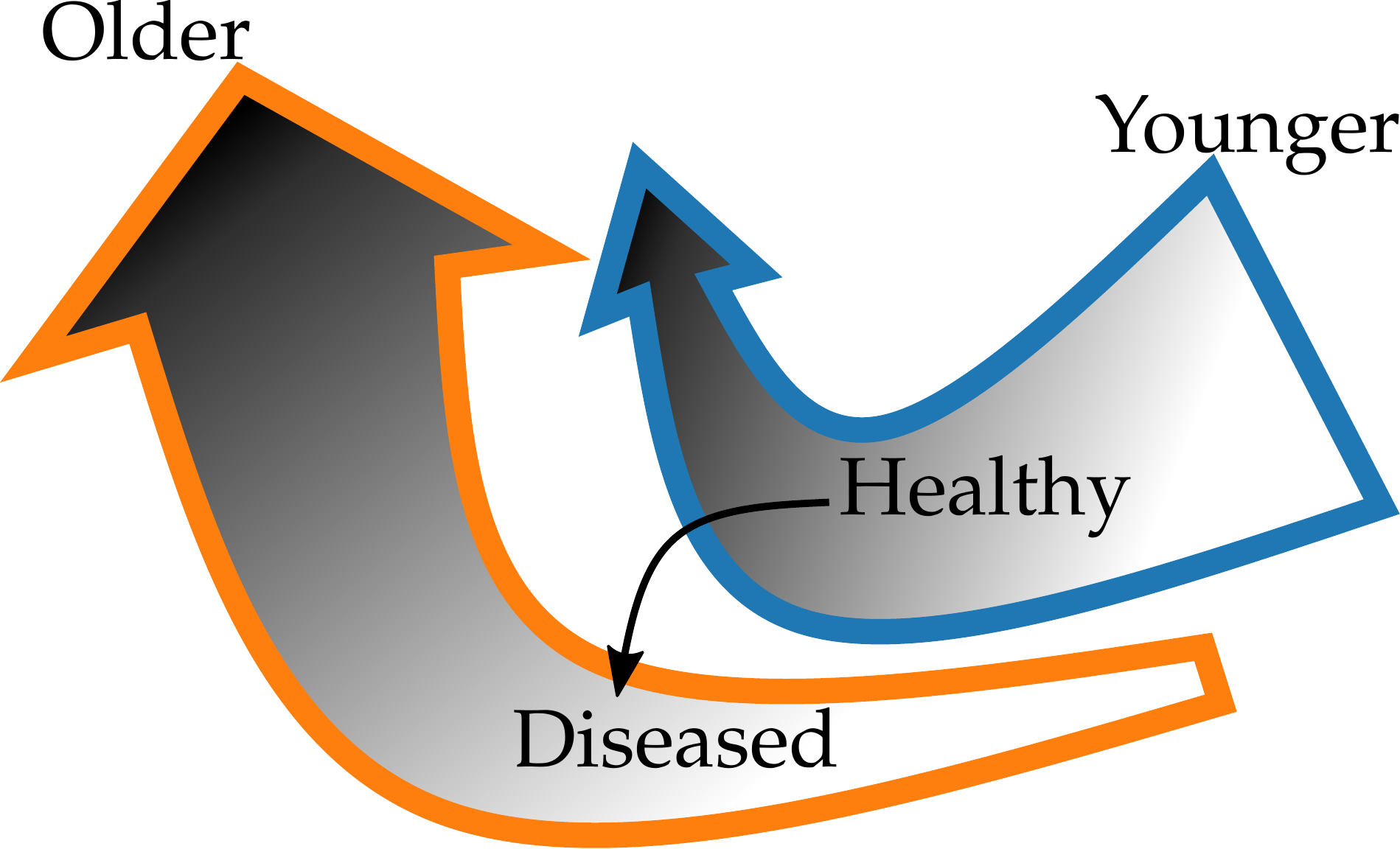}

    \hspace*{.05\linewidth}%
    \includegraphics[width=.23\textwidth]{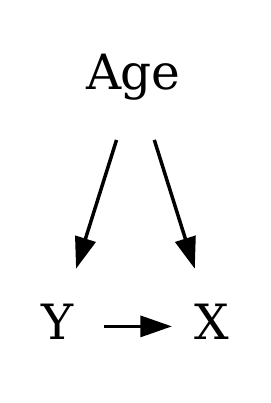}
  \end{minipage}%
  \hspace{-.2\textwidth}%
  \begin{minipage}{.69\textwidth}
    \includegraphics[width=\textwidth]{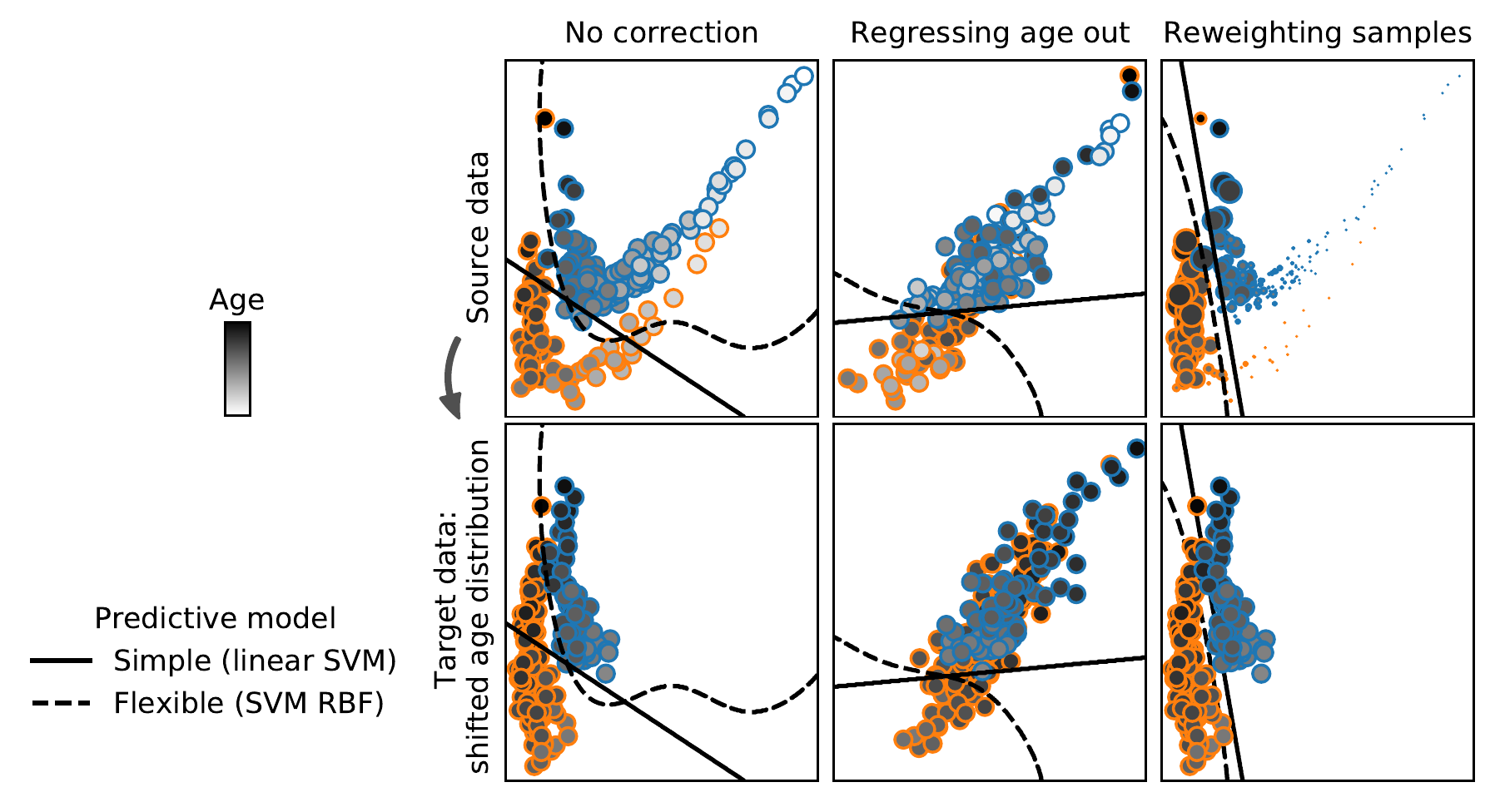}
  \end{minipage}%
\caption{\label{fig:parabolas} %
  \textbf{Classification with dataset shift -- regressing out a correlate of the
shift does not help generalization.} The task is to classify patients
(orange) from healthy controls (blue), using
2-dimensional features. Age, indicated by the shade of gray, influences
both the features and the probability of disease.
  \textbf{Left: generative process for the simulated data.} Age influences
  both the target \(Y\) and the features \(X\), and \(Y\) also has an effect on
  \(X\). Between the source and target datasets, the distribution of age
  changes.
  The two arrows point towards increasing age and represent the Healthy and
  Diseased populations, corresponding to the orange and blue clouds of
  points in the right panel.
  The grayscale gradient in the arrows represents the increasing age of
the individuals (older individuals correspond to a darker shade).
  Throughout their life, individuals can jump from the Healthy trajectory to
  the Diseased trajectory, which is slightly offset in this 2-dimensional
  feature space. As age increases, the prevalence of the
  disease increases, hence the Healthy trajectory contains more
  individuals of
  young ages (its wide end), and less at older ages (its narrow end) -- and
  vice-versa for the Diseased trajectory.
  \textbf{Right: predictive models}
  In the target data (bottom row), the age distribution is shifted:
  individuals tend to be older. Elderly are indeed often less
  likely to participate in clinical studies \citep{heiat2002representation}.
  \textbf{First column:} no correction is applied. As the situation is close to
  a covariate shift (\Cref{sec:covariate-shift}), a powerful learner (RBF-SVM)
  generalizes well to the target data. An over-constrained model -- Linear-SVM --
  generalizes poorly.
  \textbf{Second column:} wrong approach. To remove
  associations with age, features are replaced by the residuals after regressing
  them on age. This destroys the signal and results in poor performance for both
  models and datasets.
  \textbf{Third column:} Samples are weighted to
  give more importance to those more likely in the target
  distribution.  Small circles  indicate
  younger individuals, with less influence on the classifier estimation.
  This reweighting  improves prediction for the linear model on the older population.
  %
}
\end{figure*}

\paragraph{``Deconfounding'' does not correct dataset shift for
predictive models}
Dataset shift is sometimes confused with the notion of
\emph{confounding}, as both settings arise from an undesired effect in the data.
Confounding comes from \emph{causal analysis}, estimating
the effect of a \emph{treatment} --an intervention, sometimes fictional-- on an outcome.  A confounder is
a third variable --for example age, or a comorbidity-- that influences both the
treatment and the outcome. It can produce a non-causal association
between the two \citep[See][Chap. 7, for a precise definition]{hernan2020causal}.
However, the machine-learning methods we consider here capture statistical
associations, but \emph{do not target causal effects}.
Indeed, for biomarkers, the association itself is interesting, whether causal or not.
Elevated body temperature may be the consequence of a condition, but also
cause a disorder. It is a clinically useful measure in both settings.

\begin{figure*}[t!]
  \centering
  \includegraphics[width=.7\textwidth]{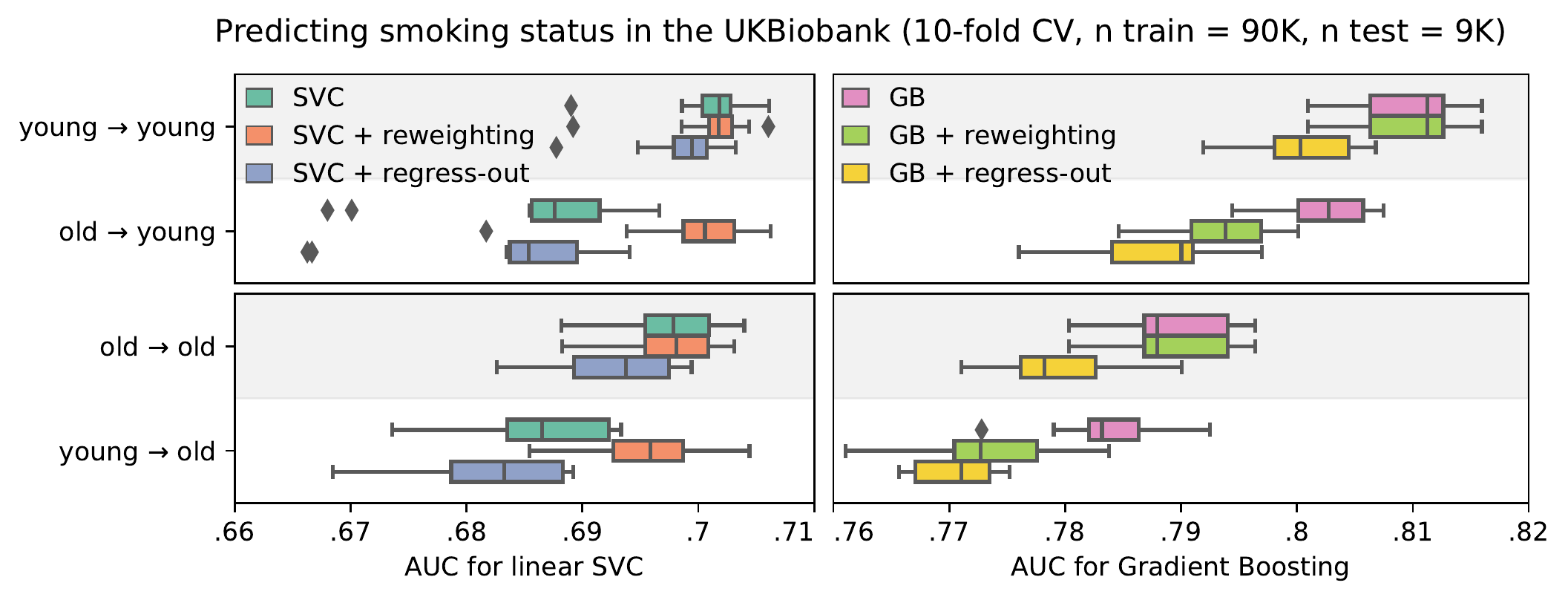}
  \caption{\label{fig:ukb-smoking} \textbf{Predicting the smoking status
of UKBiobank participants.} Different predictive models are trained on 90K UKBiobank participants and tested on 9K participants with a possibly shifted age distribution. ``young $\rightarrow$ old'' means the training set was drawn from a younger sample than the testing set. Models perform better when trained on a sample drawn from the same population as the testing set. Reweighting examples that are more likely in the test distribution (``+ reweighting'' strategy, known as Importance Weighting, \Cref{sec:importance-weighting}) alleviates the issue for the simple linear model, but is detrimental for the Gradient Boosting. Regressing  out the age (``+ regress-out'' strategy) is a bad idea and degrades prediction performance in all configurations.}
\end{figure*}

Tools for causal analysis are not all useful for prediction, 
as pointed out by seminal textbooks:
``if the
goal of the data analysis is purely predictive, no adjustment for confounding is
necessary [...] the concept of confounding does not even apply.''\citep[Sec.
18.1]{hernan2020causal}, or \citet{pearl2019seven}.
In prediction settings, applying procedures meant to adjust for confounding 
generally degrades prediction performance without solving the dataset
shift issue.
\Cref{fig:parabolas} demonstrates the detrimental effect of ``deconfounding''
on simulated data: while the target population is shifted due to a
different age distribution, removing the effect of age also removes
the separation between the two outcomes of interest.
The same behavior is visible on real epidemiologic data with age shifts,
such as predicting the
smoking status of participants in the UKBiobank study
\citep{sudlow2015uk}, as shown in \Cref{fig:ukb-smoking}.
Drawing training and testing samples with different age distributions
highlights the effect of these age shifts on prediction performance
(see \Cref{sec:ukb-experiment-details} for details on the procedure). 
For a given learner and test population, training on a different population degrades prediction.
For example, predictions on the old population are degraded when the model is trained on the young population.
A flexible model (Gradient Boosting) outperforms the linear model with or without dataset shift.
``Regressing out'' the age (as in the second column of \Cref{fig:parabolas}, ``+ regress-out'' strategy in \Cref{fig:ukb-smoking}) degrades the predictions in \emph{all} configurations.

For both illustrations on simulated and real data (\Cref{fig:parabolas}
and \ref{fig:ukb-smoking}), we also demonstrate an approach suitable for
predictive models: reweighting training examples giving more importance to those more likely in the test population.
This approach improves the predictions of the overconstrained (misspecified) linear model in the presence of dataset shift, but degrades the predictions of the powerful learner.
The non-linear model already captures the correct separation for both young and old individuals, thus reweighting examples does not bring any benefit but only increases the variance of the empirical risk.
A more detailed discussion of this approach, called \emph{importance weighting}, is provided in \Cref{sec:importance-weighting}.

\paragraph{Training examples should not be selected to be homogeneous}
To obtain valid predictive models that perform well beyond the training sample,
it is crucial to collect datasets that represent the whole population and
reflect its diversity as much as possible
\citep{kakarmath2020best,england2019artificial,o2016weapons}. Yet clinical research often
emphasizes the opposite: very
homogeneous datasets and carefully selected participants. While this may help
reduce variance and improve statistical testing, it degrades prediction
performance and fairness. In other words, the machine-learning system may perform worse for segments of the population that are under-represented in the dataset, resulting in uneven quality of care if it is deployed in clinical settings.
Therefore in \emph{predictive} settings, where the goal is 
machine-learning models that generalize well, large and diverse datasets are desirable.

\paragraph{Simpler models are not less sensitive to dataset shift}
Often, flexible models can be more robust to dataset
shifts, and thus generalize better, than linear models
\citep{storkey2009training}, as seen in
\Cref{fig:ukb-smoking,fig:parabolas}. Indeed, an over-constrained (ill-specified) model may
only fit well a restricted region of the feature space, and its performance can
degrade if the distribution of inputs changes, even if the relation to the
output stays the same (\ie when covariate shift occurs, \Cref{sec:covariate-shift}).

Dataset shift does not call for simpler models as it is not a small-sample
issue. Collecting more data from the same sources will not correct systematic dataset bias.

\section{Preferential sample selection: a common source of shift}
\label{sec:preferential-sample-selection}
In 2017, competitors in the million-dollar-prize
\href{https://www.kaggle.com/c/data-science-bowl-2017/overview}{data science
  bowl} used machine learning to predict if individuals would be diagnosed with
lung cancer within one year, based on a CT scan.
Assuming that the winning model achieves satisfying accuracy on left-out
examples from this dataset, is it ready to be deployed in hospitals? Most likely
not.
Selection criteria may 
make this dataset not 
representative of the potential lung cancer patients general population.
Selected participants verified many criteria, including being a smoker and not
having recent medical problems such as pneumonia. How would the winning
predictor perform on a more diverse population? For example, another disease
could present features that the classifier could mistakenly take for signs of lung
cancer.
Beyond explicit selection criteria, many factors such as age, ethnicity, or
socioeconomic status influence participation in biomedical studies
\citep{henrich2010most,murthy2004participation,heiat2002representation,chastain2020racial}.
Not only can these shifts reduce overall predictive performance, they can also
lead to discriminative clinical decisions for poorly represented populations
\citep{oakden2020hidden,gianfrancesco2018potential,barocas-hardt-narayanan,abbasi2020risk,cirillo2020sex}.

The examples above are instances of preferential selection, which happens when
members of the population of interest do not have equal probabilities of being
included in the source dataset: the selection \(S\) is not independent of \((X,
Y)\).
Preferential sample selection is ubiquitous and cannot always be prevented by
careful study design \citep{bareinboim2012controlling}. It is therefore a major
challenge to the identification of reliable and fair biomarkers.
Beyond preferential sample selection, there are many other sources of dataset
shifts, e.g. population changes over time, interventions such as the
introduction of new diagnostic codes in Electronic Health Records
\citep{saez2020ehrtemporalvariability}, and the use of different acquisition
devices.

\subsection{The selection mechanism influences the type of dataset shift}

The correction for a dataset shift depends on the nature of this shift, 
characterized by which and how distributions are modified \citep{storkey2009training}.
Knowledge of the mechanism producing the dataset shift 
helps formulate hypotheses about distributions that remain unchanged in the
target data \citep[Chap. 5]{scholkopf2012causal,peters2017elements}.

\begin{figure}
  \begin{minipage}{.3\textwidth}
  \begin{minipage}{\textwidth}
  \begin{minipage}{.57\textwidth}
    \includegraphics[width=\textwidth]{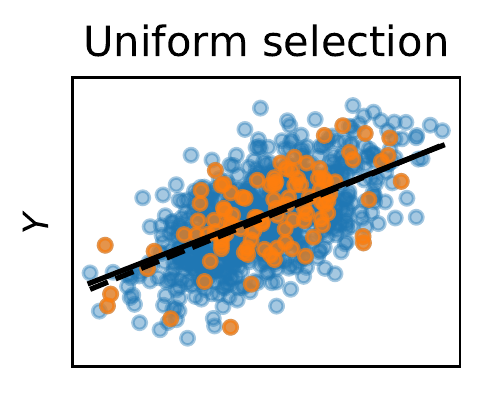}
  \end{minipage}%
  \begin{minipage}{.42\textwidth}
    \includegraphics[width=\textwidth]{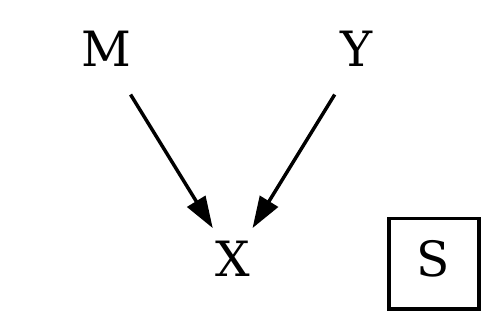}
    \centering
    \(S \indep X\,,\,Y\)
  \end{minipage}%
  \end{minipage}%

  \begin{minipage}{\textwidth}
\vspace{-7pt}
  \begin{minipage}{.57\textwidth}
    \includegraphics[width=\textwidth]{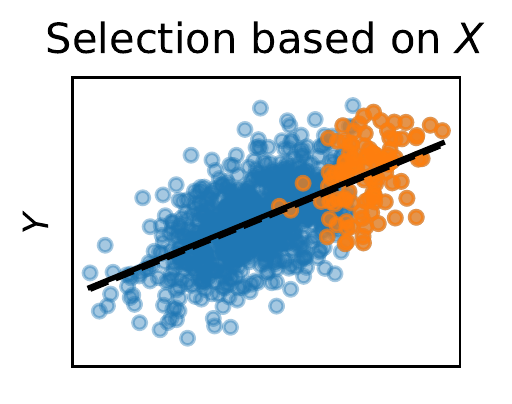}
  \end{minipage}%
  \begin{minipage}{.42\textwidth}
    \includegraphics[width=\textwidth]{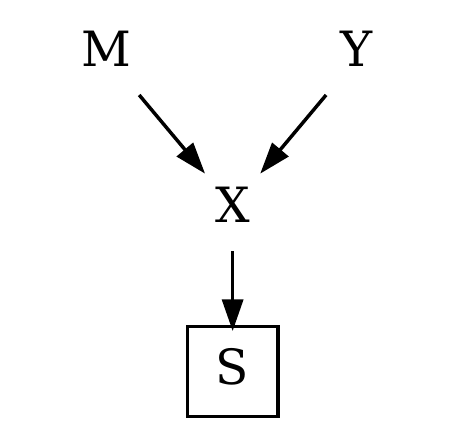}
    \centering
    \(Y \indep S \, | \, X\)
  \end{minipage}

    \begin{minipage}{\textwidth}
  \begin{minipage}{.57\textwidth}
\vspace{-7pt}
    \includegraphics[width=\textwidth]{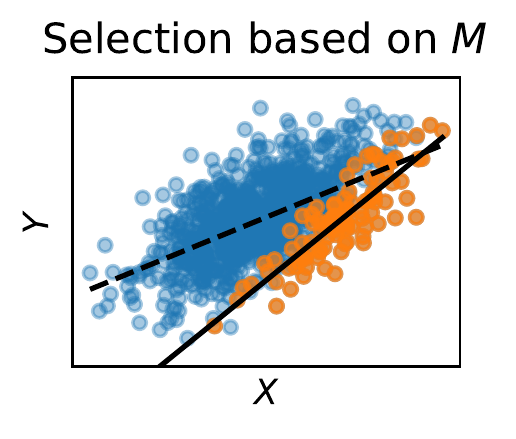}
  \end{minipage}%
  \begin{minipage}{.42\textwidth}
    \includegraphics[width=\textwidth]{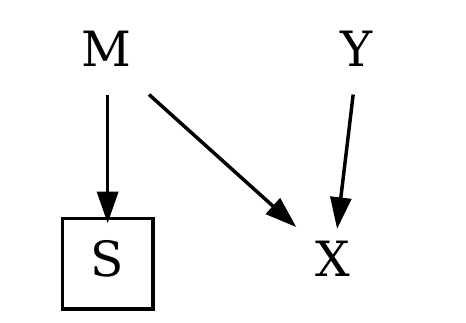}
    \centering
    \(Y \nindep S \, | \, X\)
  \end{minipage}
  \end{minipage}

  \end{minipage}
  \end{minipage}
  \caption{\textbf{Sample selection bias: three examples.}
    On the right are graphs giving conditional independence relations
    \citep{pearl2016causal}.
    \(Y\) is the lesion volume to be predicted (\ie the output).  \(M\) are the imaging
    parameters, \eg contrast agent dosage. \(X\) is the image, and depends both
    on \(Y\) and \(M\) (in this toy example \(X\) is computed as
    \(X \coloneqq Y + M + \epsilon\), where \(\epsilon\) is additive noise.
    \(S\) indicates that data is selected to enter the source dataset (orange
    points) or not (blue points).  The symbol \(\indep\) means independence
    between variables.
    Preferentially selecting samples results in a dataset shift (middle and
    bottom row). Depending on whether \(Y \indep S  \giv X\), the conditional
    distribution of \(Y \giv X\) -- here lesion volume given the image -- estimated on
    the selected data may be biased or not.
}
  \label{fig:sample-selection-bias}

\end{figure}

\Cref{fig:sample-selection-bias} illustrates this process
with a simulated example of preferential sample selection.
We consider the problem of predicting the volume \(Y\) of a tumor from
features \(X\) extracted from contrast CT images. These features can be
influenced not only by the tumor size, but also by the dosage of a contrast
agent \(M\).
The first panel of \Cref{fig:sample-selection-bias} shows a selection
of data independent of the image and tumor volume: there is no dataset shift.
In the second panel, selection depends on the CT image itself (for example images
with a low signal-to-noise ratio are discarded). As selection is independent of
the tumor volume \(Y\) given the image \(X\), the distribution of images changes but the
conditional distribution \(P(Y \giv X)\) stays the same: we face a
\emph{covariate shift} (\Cref{sec:covariate-shift}). The learned association
remains valid.
Moreover, reweighting examples to give more importance to those
less likely to be selected can improve predictions for target
data (\Cref{sec:importance-weighting}), and
it can be done with only \emph{unlabelled} examples from the target data.
In the third panel, individuals who received a low contrast agent dose are less
likely to enter the training dataset. Selection is therefore not independent of
tumor volume (the output) given the image values (the input features). Therefore
we have sample selection bias: the relation \(P(Y \giv X)\) is different in
source and target data, which will affect the performance of the prediction.
As these examples illustrate, the causal structure of the data helps identify
the type of dataset shift and what information is needed to correct it.
When such information is available, it may be possible to leverage it in order to improve robustness to dataset shift \citep[\eg][]{subbaswamy2019preventing}.

\section{Importance weighting: a generic tool against dataset shift}\label{sec:importance-weighting}
Importance weighting is a simple approach to dataset shift that applies to
many situations and can be easy to implement.
%

%
Dataset shift occurs when the joint distribution of the features and outputs is
different in the source (data used to fit the machine-learning model) and in the target data.
Informally, importance weighting consists in \emph{reweighting} the
available data to create a pseudo-sample that follows the same distribution as
the target population.

To do so, examples are reweighted by their \emph{importance weights} -- the
ratio of their likelihood in target data over source data. Examples that are 
rare in the source data but are likely in the target data are
more relevant and therefore receive higher weights.
A related approach is importance \emph{sampling} -- resampling the training data according to the importance weights.
Many statistical learning algorithms -- including Support Vector Machines,
decision trees, random forests, neural networks -- naturally
support weighting the training examples. Therefore, the challenge relies mostly
in the estimation of the appropriate sample weights and the learning algorithm
itself does not need to be modified.

\begin{figure}[h]
\centering
\includegraphics[width=.2\textwidth]{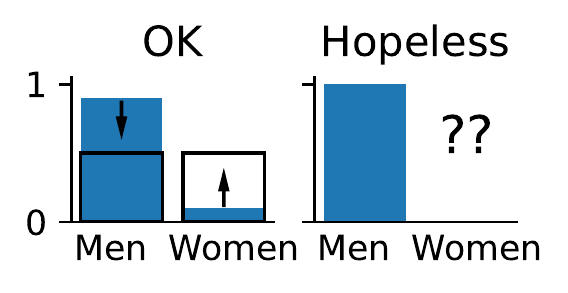}
\caption{\textbf{Dataset shifts that may or may not be compensated by
reweighting} -- \textbf{Left:} distribution of sex can be balanced by downweighting
    men and upweighting women. \textbf{Right:} women are completely missing; the
    dataset shift cannot be fixed by importance weighting.}
  \label{fig:importance-weighting-positivity}

\end{figure}

To successfully use importance weighting, no part of the target distribution
should be completely unseen.
For example, if sex (among other features) is used to predict heart failure and
the dataset only includes men, importance weighting cannot transform this
dataset and make its sex distribution similar to that of the general population
(\Cref{fig:importance-weighting-positivity}).
Conversely, the source distribution may be broader than the target distribution
(as seen for example in \Cref{fig:parabolas}).

Importance weights can also be applied to validation examples, which may produce a more accurate estimation of generalization error on target data.

Importance weighting is a well-known approach and an important body of literature focuses on its application and the estimation of importance weights.
It is illustrated on small datasets for the prediction of breast cancer in \citet{dudik2006correcting} and heart disease in \citet{kouw2019review}.
However, it cannot always be applied: some knowledge of the target distribution is required, and the source distribution must cover its support.
Moreover, importance weighting can increase the variance of the empirical
risk estimate, and thus sometimes \emph{degrades} performance -- as seen in \Cref{fig:ukb-smoking}.
It is therefore a straightforward and popular approach to consider, but not a complete solution.
It is particularly beneficial when using a simple learning model which
cannot capture the full complexity of the data, such as the linear models
in \Cref{fig:parabolas}. Indeed, simple models are often prefered in
biomedical applications because they are easy to interpret and audit.

In \Cref{sec:definition-estimation-iw}, we provide a more precise definition of
the importance weights, as well as an overview of how they can be estimated and
used.

\section{Other approaches to dataset shift}
Beyond importance weighting, many other solutions to dataset shift have been proposed.
They are typically more difficult to implement, as they require adapting or desiging new learning algorithms.
However, they may be more effective, or applicable when information about the target distribution is lacking.
We summarize a few of these approaches here.
A more systematic review can be found in \citet{kouw2019review}.
\Citet{weiss2016survey} and \citet{pan2009survey} give systematic reviews of transfer learning (a wider family of learning problems which includes dataset shift).

The most obvious solution is to do nothing -- ignoring the dataset shift.
This approach should be included as a baseline when testing on a sample of target data -- which is a prerequisite to clinical use of a biomarker \citep{storkey2009training,woo2017building}.
With flexible models, this is a strong baseline that can outperform
importance weighting, as in the right panel of \Cref{fig:ukb-smoking}.
Another approach is to learn representations--transformations of the
signal--- that are invariant to the shift \citep{achille2018emergence}.
Some deep-learning methods strive to extract features that are predictive
of the target while having similar distributions in the source and target
domains \citep[\eg][]{long2015learning}, or while preventing an adversary to distinguish source and target data \citep[``domain-adversarial'' learning, \eg][]{tzeng2017adversarial}.
When considering such methods, one must be aware of the fallacy shown in
\Cref{fig:parabolas}: making the features invariant to the effect driving the dataset shift can
remove valuable signal if this effect is not independent of the outcome
of interest.


It may also be possible to explicitly model the mapping from source to target domains, \eg by training a neural network to translate images from one modality or imaging device to another, or by relying on optimal transport \citep{courty2016optimal}.

Finally, synthetic data augmentation sometimes helps -- relying on known invariances \eg for images by applying affine transformations, resampling, \etc. or with learned generative models \citep[e.g.][]{antoniou2017data}.

\subsection{Performance heterogeneity and fairness}

It can be useful not to target a specific population, but rather
find a predictor robust to certain dataset shifts.
Distributionally robust optimization tackles this goal by
defining an ambiguity, or uncertainty set -- a set of distributions to which the target distribution might belong -- then minimizing the worse risk across all distributions in this set \citep[see][for a review]{rahimian2019distributionally}.
The uncertainty set is often chosen centered on the empirical (source) distribution for some divergence between distributions.
Popular choices for this divergence are the Wasserstein distance, \(f\)-divergences (e.g. the KL divergence) \citep{duchi2018learning}, and the Maximum Mean Discrepancy \citep{zhu2020kernel}.
If information about the target distribution is available, it can be incorportated in the definition of the uncertainty set.
An approach related to robust optimization is to strive not only to minimize
the empirical loss \(L(Y, f(X))\) but also its variance
\cite{maurer2009empirical,namkoong2017variance}.


%
It is also useful to assess model performance across values of demographic
variables such as age, socioeconomic status or ethnicity. Indeed,
a good overall prediction performance can be achieved despite a poor
performance on a minority group. 
Ensuring that a
predictor performs well for all subpopulations reduces sensitivity to potential
shifts in demographics and is essential to ensure fairness
\citep{abbasi2020risk}.
For instance, there is a risk that machine-learning analysis of dermoscopic images under-diagnoses malignant moles on skin tones that are typically under-represented in the training set \cite{adamson2018machine}.
Fairness is especially relevant when the model output could be used to grant
access to some treatment.
As similar issues arise in many applications of
machine learning, there is a growing literature on fairness
\citep[see \eg][for an overview]{barocas-hardt-narayanan}.
For instance, 
\citet{duchi2018learning} show that distributionally robust optimization can
help performance on under-represented subpopulations.

\subsection{Multi-site datasets}

Often datasets are collected across several sites or hospitals, or with
different measurement devices. This heterogeneity
provides an opportunity to train models that generalize to
unseen sites or devices. Some studies attempt to remove site effects by
regressing all features on the site indicator variable. For the same reasons
that regressing out age is detrimental in \Cref{fig:parabolas}, this
strategy often gives worse generalization across sites.

Data harmonization, such as compensating differences across measurement devices, is crucial, but remains very difficult and cannot correct these differences perfectly \citep{glocker2019machine}.
Removing too much inter-site variance can lead to loss of informative
signal. Rather, it is important to model it well, accounting for the two
sources of variance, across participants and across sites. A good model
strives to yield good results on all sites. One solution is to adapt
ideas from robust optimization: on data drawn from different
distributions (\eg from several sites), \citet{krueger2020out} show the
benefits of
minimizing the empirical risk on the worse site or adding penalties on the
variance of the loss across sites.

Measures of prediction performance should aggregate scores at
the site level (not pooling all individuals), and check the variance across
sites and the performance on the worse site. Cross-validation schemes should
hold out entire sites \citep{woo2017building,little2017using}.
%




\section{Special cases of dataset shift}

Categorizing dataset shift helps finding the best approach to tackle it
\cite{storkey2009training,moreno2012unifying}.
We summarize two frequently-met scenarios that are easier to handle than the general case and can call for different adjustments: covariate shift (\Cref{sec:covariate-shift}) and prior probability shift (\Cref{sec:prior-probability-shift}).

\subsection{Covariate shift}
\label{sec:covariate-shift}

\begin{figure}
\centering
\includegraphics[width=.7\linewidth]{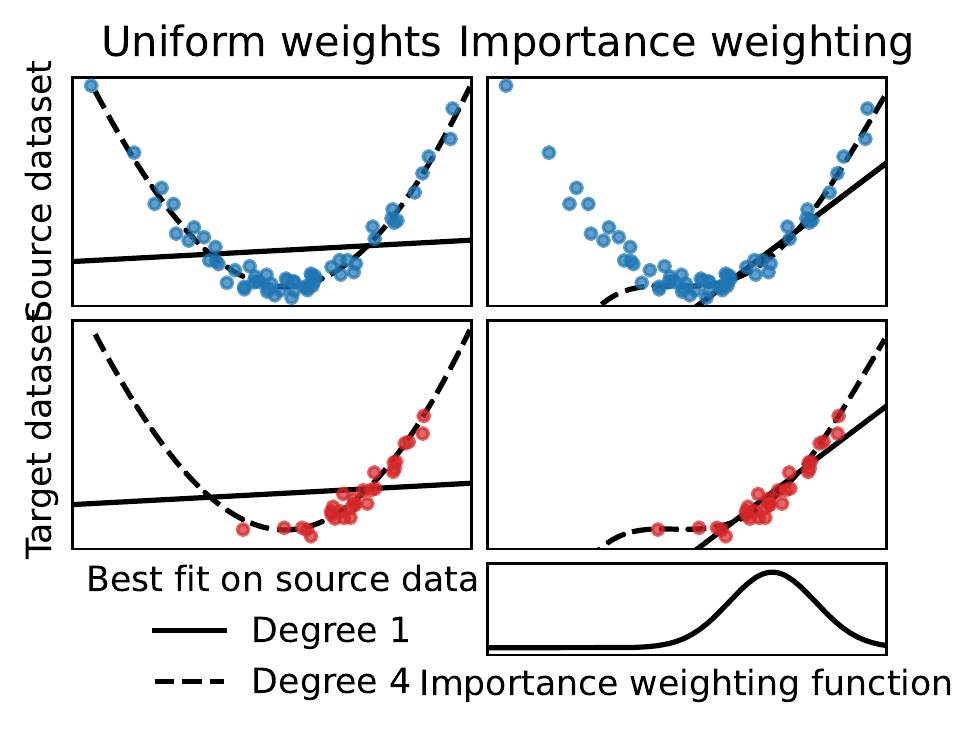}
\caption{\label{fig:covariate-shift}\textbf{Covariate shift:} \(\Prob(Y \giv
  X)\) stays the same but the feature space is sampled differently in the source
  and target datasets. A powerful learner may generalize well as \(\Prob(Y \giv
  X)\) is correctly captured \citep{storkey2009training}. Thus the polynomial
  fit of degree 4 performs well on the new dataset. However, an overconstrained
  learner such as the linear fit can benefit from reweighting training examples
  to give more importance to the most relevant region of the feature space.}
\end{figure}

Covariate shift occurs when the marginal distribution of \(X\) changes between
the source and target datasets (i.e. \( p_t(x) \neq p_s(x) \)), but \(P(Y \giv X)\) stays the same.
This happens for example in the second scenario in
\Cref{fig:sample-selection-bias}, where sample selection based on \(X\) (but not
\(Y\)) changes the distribution of the inputs.
If the model is correctly specified, an estimator trained with uniform weights
will lead to optimal predictions given sufficient training data
\citep[prediction consistency][Lemma 4]{shimodaira2000improving}.
However the usual (unweighted) estimator is not consistent for an
over-constrained (misspecified) model.
Indeed, a over-constrained model may be able to fit the data well only in some
regions of the input feature space (\Cref{fig:parabolas}). In this case reweighting training examples (\Cref{sec:importance-weighting}) to give
more importance to those that are more representative of the target data is
beneficial \citep{storkey2009training,scholkopf2012causal}.
\Cref{fig:covariate-shift} illustrates covariate shift.
%

\subsection{Prior probability shift}
\label{sec:prior-probability-shift}

\begin{figure}
  \centering
  \includegraphics[width=.5\linewidth]{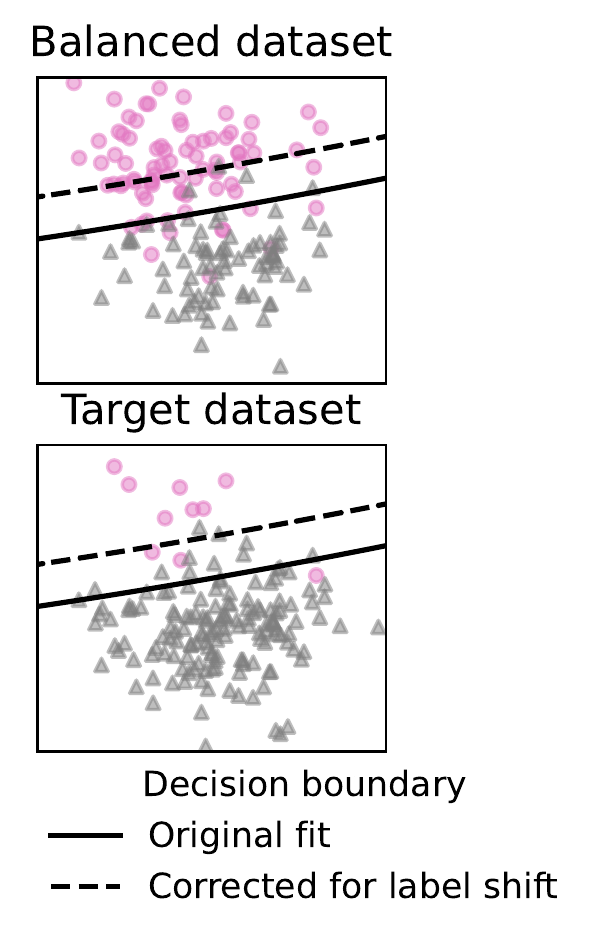}
  \caption{
    \textbf{Prior probability shift:} when \(P(Y)\) changes but \(P(X
    \giv Y)\) stays the same. This can happen for example when participants are
    selected based on \(Y\) -- possibly to have a dataset with a balanced number
    of patients and healthy participants: \(X \leftarrow Y \rightarrow \text{\fbox{$S$}}\).
    When the prior probability (marginal distribution of \(Y\)) in the
    target population is known, this is easily corrected by applying Bayes' rule.
    The output \(Y\) is typically low-dimensional and discrete
    (often it is a single binary value), so \(P(Y)\) can often be estimated
    precisely from few examples.}
  \label{fig:label-shift-scatter}
\end{figure}

Another relatively simple case of dataset shift is \emph{prior probability shift}.
With prior probability shift (\aka label shift or target shift), the
distribution of \(Y\) changes but not \(P(X \giv Y)\).
This happens for example when disease prevalence changes in the target population but manifests itself in the same way.
Even more frequently, prior probability shift arises when one rare class is over-represented in the training data so that the dataset is more balanced, as when extracting a biomarker from a case-control cohort, or when the dataset is resampled as a strategy to handle the \emph{class imbalance} problem \citep{he2009learning}.
Prior probability shift can be corrected without extracting a new biomarker, simply by adjusting a model's predicted probabilities using Bayes' rule \citep[as noted for example in][]{storkey2009training,scholkopf2012causal}.
When the classes are well separated, the effect of this correction may be small, \ie the uncorrected classifier may generalize well without correction.
\Cref{fig:label-shift-scatter} illustrates prior probability shift.

\section{Conclusion}
Ideally, machine learning biomarkers would be designed and trained using
datasets carefully collected to be representative of the
targeted population -- as in \citet{liu2020sensitive}.
To be trusted, biomarkers ultimately need to be evaluated rigorously on one or several
independent and representative samples.
However, such data collection is expensive. It is therefore useful to
exploit existing datasets in an opportunistic way as much as possible in the
early stages of biomarker development.
When doing so, correctly accounting for dataset shift can prevent wasting
important resources on machine-learning predictors that have little chance of
performing well outside of one particular dataset.

We gave an overview of importance weighting, a simple tool against dataset
shift.
Importance weighting needs a clear definition of the targeted population and
access to a diverse training dataset. When this is not possible,
distributionally robust optimization may be promising alternative, though it
is a more recent approach and more difficult to implement.
Despite much work and progress, dataset shift remains a difficult problem.
Characterizing its impact and the effectiveness of existing solutions for biomarker discovery will be important for machine learning models to become more reliable in healthcare applications.

We conclude with the following recommendations:
\begin{itemize}
  \item be aware of the dataset shift problem and the difficulty of out-of-dataset generalization. Do not treat cross-validation scores on one dataset as a guarantee that a model will perform well on clinical data.
  \item collect diverse, representative data.
  \item use powerful machine-learning models and large datasets.
  \item consider using importance weighting to correct biases in the data
collection, especially if the learning model may be over-constrained (\eg when using a linear model).
  \item look for associations between prediction performance and demographic variables in the validation set to detect potential generalization or fairness issues.
  \item \emph{do not} remove confounding signal in a predictive setting.
\end{itemize}
These recommendations should help designing fair biomarkers and their efficient application on new cohorts.

\paragraph{Author contributions}
Jérôme Dockès, Gaël Varoquaux and Jean-Baptiste Poline participated in
conception, literature search, data interpretation, and editing the manuscript.
Jérôme Dockès wrote the software and drafted the manuscript. Both Gaël Varoquaux
and Jean-Baptiste Poline contributed equally to this work (as last authors).

\paragraph{Competing interests statement}
The authors declare that there are no competing interests.

\paragraph{Software and data availability}
The source files used to create this publication can be found in this repository:
\url{https://github.com/neurodatascience/dataset_shift_biomarkers}.
UKBiobank data can be obtained from \url{https://www.ukbiobank.ac.uk}.


\bibliography{biblio}
\appendix
\section{Definition and estimation of importance weights}
\label{sec:definition-estimation-iw}
We will implicitly assume that all the random variables we consider admit
densities and denote \(p_s\) and \(p_t\) the density of the joint distribution of \((X, Y)\) applied
to the source and target populations respectively.
%
If the support of \(p_t\) is included in that of \(p_s\) (meaning that \(p_s > 0
\) wherever \( p_t > 0\)), we have:
%
\begin{equation}
\E_{\text{source}}[\,L(Y,\,f(X))\,] = \E_{\text{target}}\left[\,\frac{p_t(X,\,Y)}{p_s(X,\,Y)}\;L(Y,\,f(X))\,\right] \; ,
\end{equation}
%
where \(L\) is the cost function and \(f\) is a prediction function, \(\E_{\text{source}}\) (resp. \(\E_{\text{target}}\)) the expectation on the source (resp. target) data.
The risk (on target data) can therefore be computed as an expectation on the
source distribution where the loss function is reweighted by the
\emph{importance weights}:
\begin{equation}\label{eq:importance-weight-definition}
w(x,y) = \frac{p_t(x,y)}{p_s(x,y)} \; .
\end{equation}
If \(\hat{w}\) are empirical estimates  of the importance weights \(w\), it is possible to compute the reweighted empirical risk:
\begin{equation}
\hat{R}_{\hat{w}}(f) = \sum_{i=1}^n \hat{w}(x_i, y_i)\,L(y_i, f(x_i)) \; .
\end{equation}
%
%
%

Rather than being weighted, examples can also be resampled with importance or rejection
sampling \citep{zadrozny2003cost,zadrozny2004learning}. Importances can also be
taken into account for model selection -- for example in
\citet{sugiyama2007covariate} examples of the test set are also reweighted when
computing cross-validation scores.
\Citet{cortes2008sample} study how errors in the estimation of the weights
affect the prediction performance.

\subsection{Preferential Sample selection and Inverse Probability weighting}

In the case of preferential sample selection
(\Cref{sec:preferential-sample-selection}), the condition that requires for the
support of \(p_t\) to be included in the support of \(p_s\) translates to a
requirement that all individuals have a non-zero probability of being selected:
\(P(S=1 \giv x,y) > 0\) for all \((x, y)\) in the support of \(p_t\).
When this is verified, by applying Bayes' rule the definition of importance
weights in \Cref{eq:importance-weight-definition} can be reformulated
\citep[see][Sec. 2.3]{cortes2008sample}:
\begin{equation}\label{eq:inverse-probability-weight-definition}
  w(x, y) = \frac{P(S=1)}{P(S=1 \giv X=x,Y=y)} \;
\end{equation}
These weights are sometimes called Inverse Probability weights
\citep{hernan2004structural} or Inverse Propensity scores
\citep{austin2011introduction}. Training examples that had a low probability of
being selected receive higher weights, because they have to account for similar
individuals who were not selected.

\subsection{Computing importance weights}

In practice \(p_t(x, y)\), which is the joint density of \((X,
Y)\) in the target data, is not known. However, it is not needed for the estimation of \(p_t / p_s\).
More efficient estimation hinges on two observations: estimation
of both densities separately is not necessary to estimate their ratio, and
variables that have the same distribution in source and target data can be factored out.

Here we describe methods that estimate the true importance weights \(p_t /
p_s\), but we point out that reweighting the training examples reduces the bias
of the empirical risk but increases the variance of the estimated model
parameters.
%
%
Even when the importances are perfectly known, it can therefore be beneficial to
regularize the weights \citep{shimodaira2000improving}.

\subsubsection{Computing importance weights does not require distributions
  densities estimation}\label{sec:no-density-estimation}

Importance weights can be computed by modelling separately \(p_s\) and \(p_t\)
and then computing their ratio \citep[Sec. 4.1]{sugiyama2012machine}.
However, distribution density estimation is notoriously difficult; non-parametric methods
suffer from the curse of dimensionality and parametric methods depend heavily on
the correct specification of a parametric form.

But estimating both densities is more information than is needed to compute
the sample weights.
Instead, one can directly optimize importance weights in order to make the
reweighted sample similar to the target distribution, by matching moments
\citep{sun2016return} or mean embeddings
\citep{huang2007correcting,zhang2013domain}, minimizing the KL-divergence
\citep{sugiyama2008direct}, solving a least-squares estimation problem
\citep{kanamori2009least} or with optimal transport \citep{courty2016optimal}.

Alternatively, a discriminative model can be trained to distinguish source and target
examples. In the specific case of preferential sample selection, this means estimating
directly the probability of selection \(P(S=1)\) (cf
\Cref{eq:inverse-probability-weight-definition}).
In general, the shift is not always due to selection: the source data is not
necessarily obtained by subsampling the target population. In this case we
denote \(T = 1\) if an individual comes from the target data and \(T = 0\) if it
comes from the source data.
Then, a classifier can be trained to predict from which dataset (source
or target) a sample is drawn, and the importance weights obtained from the
predicted probabilities \citep[Sec. 4.3]{sugiyama2012machine}:
\begin{equation}
  w(x, y) = \frac{P(T=1 \giv X=x, Y=y)\,P(T=0)}{P(T=0 \giv X=x, Y=y)\,P(T=1)} \;,
\end{equation}

The classifier must be calibrated (i.e. produce accurate
probability estimates, not only a correct decision), see
\citet{niculescu2005predicting}.
Note that constant factors such as \(P(T=0) / P(T=1)\) usually do not matter and are easy
to estimate if needed.
This discriminative approach is effective because the distribution of \((T \giv
X=x, Y=y)\) is much easier to estimate than the distribution of \((X, Y \giv
T=t)\) : \(T\) is a single binary variable whereas \((X, Y)\) is
high-dimensional and often continuous.

The classifier does not need to distinguish source and target examples with high
accuracy. In the ideal situation of no dataset shift, the classifier will
perform at chance level. On the contrary, a high accuracy means that there is
little overlap between the source and target distributions and the model
will probably not generalize well.

\subsubsection{What distributions differ in source and target data?}

When computing importance weights, it is possible to exploit prior knowledge that some distributions are left
unchanged in the target data.
For example,
\begin{equation}
  \frac{p_t(x, y)}{p_s(x, y)} = \frac{p_t(y \giv x)\,p_t(x)}{p_s(y \giv x)\,p_s(x)} \; .
\end{equation}

Imagine that the marginal distribution of input \(X\) differs in source and
target data, but the conditional distribution of the output \(Y\) given the input
stays the same: \(p_t(x) \neq p_s(x)\) but \(p_t(y \giv x) = p_s(y \giv x)\) (a
setting known as \emph{covariate shift}). Then, the importance weights
simplify to
\begin{equation}
  w(x, y) = \frac{p_t(x)}{p_s(x)} \; .
\end{equation}
In this case, importance weights can be estimated using only unlabelled examples
(individuals for whom \(Y\) is unkown) from the target distribution.

Often, the variables that influence selection (\eg demographic variables such as
age) are lower-dimensional than the full features (\eg high-dimensional images),
and dataset shift can be corrected with limited information on the target
distribution, with importance weights or otherwise.
Moreover, even if additional information \(Z\) that predicts
selection but is independent of \((X, Y)\) is available, it should \emph{not} be used to
compute the importance weights. Indeed, this would only increase the weights'
variance without reducing the bias due to the dataset shift \citep[Sec.
15.5]{hernan2020causal}.

\section{Tobacco smoking prediction in the UKBiobank}
\label{sec:ukb-experiment-details}
We consider predicting the smoking status of participants in the UKBiobank study to illustrate the effect of dataset shift on prediction performance.

6,000 participants are used in a preliminary step to identify the 29 most relevant predictive features (listed in \cref{sec:ukb-features}), by cross-validating a gradient boosting model and computing permutation feature importances.
We then draw two samples of 100K individuals from the rest of the dataset, that have different age distributions.
In the young sample, 90\% of individuals come from the youngest 20\% of the dataset, and the remaining 10\% are sampled from  the oldest 20\% of the dataset.
In the old sample, these proportions are reversed.
We then perform 10-fold cross validation.
For each fold, both the training and testing set can be drawn from either the young or the old population, resulting in four tasks on which several machine-learning estimators are evaluated.
We use this experiment to compare 2 machine-learning models: a simple one -- regularized linear Support Vector Classifier, and a flexible one -- Gradient Boosting.
For each classifier, 3 strategies are considered to handle the dataset shift: (i) baseline -- the generic algorithm without modifications, (ii) Importance Weighting (\Cref{sec:importance-weighting}), and (iii) the (unfortunately popular) non-solution: ``regressing out the confounder'' -- regressing the predictive features on the age and using the residuals as inputs to the classifier.

The results are similar to those seen with simulated data in \Cref{fig:parabolas}.
For a given learner and test population, training on a different population degrades the prediction score.
For example, if the learner is to be tested on the young population, it performs best when trained on the young population.
Gradient Boosting vastly outperforms the linear model in all configurations.
Regressing out the age always degrades the prediction; it is always worse than the unmodified baseline, whether a dataset shift is present or not.
Finally, Importance Weighting (\Cref{sec:importance-weighting})\todo[inline]{Gael: The narrative is awkward, because we
are discussing importance weighting before introducing it. I wonder if we
should revisit the order of the sections? Jerome: note that would mean
moving figure 1 towards the end of the paper. I changed ``IW'' to
``reweighting'' so we only use the intuiton but not the term
``importance'' until IW is introduced} improves the predictions of the
over-constrained (misspecified) linear model in the presence of dataset shift, but degrades the prediction of the powerful learner used in this experiment.
This is due to the fact that the Gradient Boosting already captures the correct separation for both young and old individuals, and therefore Importance Weighting does not bring any benefit but only reduces the effective training sample size by increasing the variance of the empirical risk.

\subsection{Features used for tobacco smoking status prediction}
\label{sec:ukb-features}
The 30 most important features were identified in a preliminary experiment with 6,000 participants (that were not used in the subsequent analysis).
One of these features, ``Date F17 first reported (mental and behavioural disorders due to use of tobacco)'', was deemed trivial -- too informative, as it directly implies that the participant does smoke tobacco, and removed.
The remaining 29 features were used for the experiment described in \Cref{sec:misconceptions}.

\begin{itemize}
\item Forced expiratory volume in 1-second (FEV1), predicted percentage
\item Lifetime number of sexual partners
\item Age first had sexual intercourse
\item Age when last took cannabis
\item Ever taken cannabis
\item Forced expiratory volume in 1-second (FEV1), predicted
\item Acceptability of each blow result
\item Mouth/teeth dental problems
\item Coffee intake
\item FEV1/ FVC ratio Z-score
\item Alcohol intake frequency.
\item Date J44 first reported (other chronic obstructive pulmonary disease)
\item Former alcohol drinker
\item Average weekly spirits intake
\item Year of birth
\item Acceptability of each blow result
\item Date of chronic obstructive pulmonary disease report
\item Leisure/social activities
\item Morning/evening person (chronotype)
\item Mean sphered cell volume
\item Lymphocyte count
\item Townsend deprivation index at recruitment
\item Age hay fever, rhinitis or eczema diagnosed
\item Age started oral contraceptive pill
\item White blood cell (leukocyte) count
\item Age completed full time education
\item Age at recruitment
\item Workplace had a lot of cigarette smoke from other people smoking
\item Wheeze or whistling in the chest in last year
\end{itemize}

\section{Glossary}

Here we provide a summary of some terms and notations used in the paper.

\begin{description}
  \item [Target population] the population on which the biomarker
    (machine-learning model) will be applied.
  \item [Source population] the population from which the sample used to train
    the machine-learning model is drawn.
  \item [Selection] in the case that source data are drawn (with non-uniform
    probabilities) from the target population, we denote by $S = 1$ the fact
    that an individual is selected to enter the source data (\eg to participate
    in a medical study).
  \item [Provenance of an individual] when samples from
    both the source and the target populations (\eg
    \Cref{sec:no-density-estimation}) are available, we also denote $T=1$ if an individual
    comes from the target population and $T=0$ if they come from the source
    population.
  \item [Confounding] in \emph{causal inference}, when estimating the effect of a treatment on an outcome, confounding occurs if a third variable (\eg age, a comorbidity, the seriousness of a condition) influences both the treatment and the outcome, possibly producing a spurious statistical association between the two.
    This notion is not directly relevant to dataset shift, and we mention it only to insist that it is a different problem.
    See \citet{hernan2020causal}, Chap. 7, for a more precise definition.
  \item [Domain adaptation] the task of designing machine-learning methods that are resilient to dataset shift -- essentially a synonym for dataset shift, \ie another useful search term for readers looking for further information on this problem.
\end{description}


\end{document}